%% file: learning_to_grasp_on_the_moon.tex
\documentclass[letterpaper, 10 pt, conference]{ieeeconf}
\IEEEoverridecommandlockouts
\input{_style/dependencies_and_style.tex}

\graphicspath{{./graphics/}}

\begin{document}
\input{frontmatter/title_and_author.tex}

\begin{abstract}
	\input{frontmatter/abstract.tex}
\end{abstract}

\input{content/introduction.tex}
\input{content/related_work.tex}
\input{content/methods.tex}
\input{content/results.tex}
\input{content/discussion_and_conclusion.tex}

\input{learning_to_grasp_on_the_moon.bbl}

\end{document}

%% file: _style/dependencies_and_style.tex
\usepackage{amsmath,amssymb,amsfonts}
\usepackage{graphicx}
\usepackage{xcolor}
\usepackage[hyphens]{url}
\usepackage[hidelinks,pdftex]{hyperref}
\usepackage[T1]{fontenc}
\usepackage[all]{hypcap}

%% file: frontmatter/title_and_author.tex
\title{\bf Learning to Grasp on the Moon from 3D Octree Observations\\with Deep Reinforcement Learning}

\author{Andrej~Orsula\textsuperscript{1} \and Simon~Bøgh\textsuperscript{2} \and Miguel~Olivares-Mendez\textsuperscript{1} \and Carol~Martinez\textsuperscript{1}}

\maketitle
\thispagestyle{empty}
\pagestyle{empty}

\footnotetext[1]{Space Robotics Research Group~(SpaceR), Interdisciplinary Centre for Security, Reliability and Trust~(SnT), University of Luxembourg, Luxembourg. {\tt \{andrej.orsula}, {\tt miguel.olivaresmendez}, {\tt carol.martinezluna\}@uni.lu}}
\footnotetext[2]{Robotics \& Automation Group, Department of Materials and Production, Aalborg University, Denmark. {\tt sb@mp.aau.dk}}

\hypersetup{%
	pdfinfo ={%
			Title={Learning to Grasp on the Moon from 3D Octree Observations with Deep Reinforcement Learning},%
			Subject={2022 IEEE/RSJ International Conference on Intelligent Robots and Systems (IROS)},%
			Author={Andrej Orsula, Simon Bøgh, Miguel Olivares-Mendez and Carol Martinez},%
			Creator={LaTeX},%
			Keywords={Space Robotics and Automation; Reinforcement Learning; Deep Learning in Grasping and Manipulation},%
		}%
}%

%% file: frontmatter/abstract.tex
Extraterrestrial rovers with a general-purpose robotic arm have many potential applications in lunar and planetary exploration. Introducing autonomy into such systems is desirable for increasing the time that rovers can spend gathering scientific data and collecting samples. This work investigates the applicability of deep reinforcement learning for vision-based robotic grasping of objects on the Moon. A novel simulation environment with procedurally-generated datasets is created to train agents under challenging conditions in unstructured scenes with uneven terrain and harsh illumination. A model-free off-policy actor-critic algorithm is then employed for end-to-end learning of a policy that directly maps compact octree observations to continuous actions in Cartesian space. Experimental evaluation indicates that 3D data representations enable more effective learning of manipulation skills when compared to traditionally used image-based observations. Domain randomization improves the generalization of learned policies to novel scenes with previously unseen objects and different illumination conditions. To this end, we demonstrate zero-shot sim-to-real transfer by evaluating trained agents on a real robot in a Moon-analogue facility.
\textit{The source code and datasets are available at \href{https://github.com/AndrejOrsula/drl_grasping}{https://github.com/AndrejOrsula/drl\_grasping}.}

%% file: content/introduction.tex
\section{Introduction}\label{sec:introduction}

Planetary exploration aims to provide scientific insights to advance our knowledge about other planets, their geology and available resources. Analysis of samples plays a key role in this endeavor, where extraterrestrial robotic systems are instrumental in their acquisition for either in-situ analysis or sample return~\cite{zhang_progress_2019}. Reports of these findings are also necessary for future in-situ resource utilization~\cite{moses_frontier_2016} to enable extraction of Hydrogen and Oxygen for localized production of rocket propellant and operation of life support systems. In this way, the required payload during the initial launch from Earth would be significantly reduced while decreasing the dependency on additional resupply missions. There is an increasing effort put into sample return missions that could provide this data. Lunar material was recently returned to Earth by the Chang'e~5 mission~\cite{liu_landing_2021}, and NASA has selected companies to collect Moon rocks towards the progress of the Artemis program~\cite{schierholz_nasa_2020}. Mars Sample Return is another proposed mission, where an ESA rover is planned to fetch samples that are being collected by the NASA rover Perseverance~\cite{muirhead_mars_2019}. Unfortunately, communication delay causes remote teleoperation to be inefficient, which reduces the amount of scientific data that rovers can gather throughout their mission. Therefore, autonomy for extraterrestrial rovers becomes essential as the complexity of missions increases.

\begin{figure}[t]
	\vspace{2.057mm}
	\centering
	\includegraphics[width=1.0\linewidth]{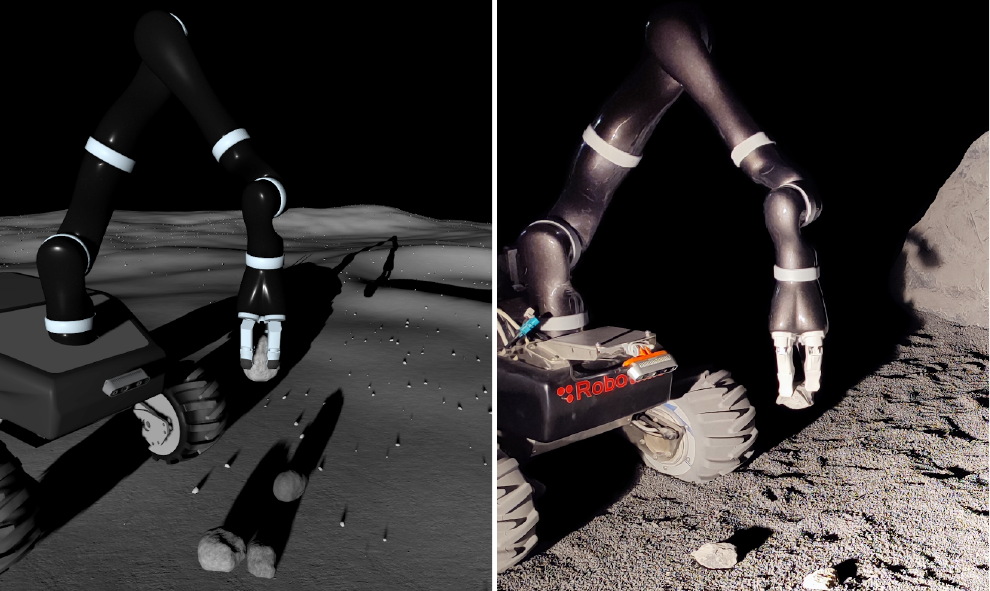}
	\caption{Our agents learn vision-based robotic grasping under challenging conditions in a heavily randomized simulation environment. Their performance is then evaluated on a real robot inside a Moon-analogue facility.}
	\label{fig:front_page_figure}
	\vspace{-1.0\baselineskip}
\end{figure}

Rovers equipped with robotic arms have many potential applications in extraterrestrial environments. Besides manipulating science instruments to closely analyze areas of interest, these rovers could also perform assembly and maintenance tasks by interacting with various tools and technical equipment. Many subroutines involved in such tasks require an object or a tool to be firmly grasped prior to performing them. Therefore, robotic grasping is a fundamental skill that is essential for flexible mobile manipulation. In order to achieve this flexibility, rovers are required to grasp a diversity of objects that can differ in their geometry, appearance and mechanical properties.

We present an approach for applying end-to-end deep reinforcement learning to the task of vision-based robotic grasping in lunar environments. The primary focus of this paper is to learn end-to-end policies for robotic grasping under challenging conditions of the Moon, i.e.~unstructured scenes with uneven terrain, diverse rocks and harsh illumination. As the training of agents directly in extraterrestrial conditions is unfeasible due to the high cost and safety requirements of space robotic systems, our goal is to employ simulations with the aim of transferring learned policies to a real robot.

The main contributions of this work are as follows:
\begin{itemize}
	\item A simulation environment of the Moon, which enables learning of mobile manipulation skills that are transferable to the real-world domain due to its realistic physics, physically-based rendering and extensive use of domain randomization with procedurally-generated datasets for simulating the wide variety of lunar conditions.
	\item A novel approach for utilizing 3D octree visual observations with multi-channel features for end-to-end deep reinforcement learning. Octrees are used to efficiently represent the 3D scene, while an octree-based convolutional neural network is applied to extract abstract features that allow agents to generalize over spatial positions and orientations.
	\item A demonstration of learning robotic grasping inside a realistic simulation environment of the Moon and subsequent zero-shot sim-to-real transfer to a real robot inside a Moon-analogue facility.
\end{itemize}

%% file: content/related_work.tex
\section{Related Work}\label{sec:related-work}

Extraterrestrial environments pose many challenges to both the hardware and software of robotic systems. Even though rovers deployed to Moon and Mars can travel long distances, their actions are controlled remotely by operators from Earth due to their limited autonomous capabilities~\cite{grotzinger_mars_2012}. Current manipulation for planetary exploration relies on using calibrated systems to achieve the required precision. However, their control is based on the manual selection of target waypoints that are acquired from stereo vision~\cite{nickels_vision_2010}. Due to limitations caused by communication delays, an effort has been put into increasing the autonomy of extraterrestrial rovers. One example of an autonomous rover featuring a robotic arm with a mechanical gripper was presented in~\cite{schuster_lru_2016}. With high-level commands, this rover was able to autonomously pick and assemble known objects by planning motions to predefined configurations relative to their pose estimate. More complex mobile manipulation tasks with the same rover were later conducted in a Moon-analogue environment~\cite{lehner_mobile_2018}, albeit the mechanical gripper was replaced with a docking interface to reduce the requirement for high precision positioning by increasing misalignment tolerance. Our work similarly investigates manipulation in lunar environments but focuses on grasping previously unknown objects with a general-purpose mechanical gripper to enable application versatility, despite increasing the task difficulty.

The mobility of rovers with robotic arms extends their reachable workspace, which in turn enhances their capabilities for interacting with the environment. However, the complexity of extraterrestrial environments poses several difficulties due to their unstructured nature and uncertainties caused by limited knowledge of surroundings with imperfect sensory perception. Analytical approaches for robotic grasping lack the required generalization even for terrestrial applications~\cite{sahbani_overview_2012}, despite their effectiveness for task-specific problems~\cite{schuster_lru_2016,lehner_mobile_2018}. Contrary to this, supervised learning provides a way to learn grasp synthesis empirically from labeled datasets. However, this approach requires a large volume of data in order to achieve the desired level of generalization~\cite{mahler_dex-net_2017}. Reinforcement learning enables acquiring policies for sequential decision-making problems by interacting with the environment in a self-supervised manner. Recent research has also applied this method in the space robotics domain for tasks such as planetary soft landing~\cite{xu_deep_2021} and rover path planning~\cite{jin_value_2021}. The application of deep reinforcement learning, i.e. a combination of deep learning and reinforcement learning, has been extensively explored for terrestrial robotic grasping in the last few years~\cite{kroemer_review_2021}. Many of these contributions focus on the final performance using a single object~\cite{popov_data-efficient_2017} or several different objects with simple geometry~\cite{tobin_domain_2017,zeng_learning_2018}. More recent work strives to increase this variety by training on objects with more complex geometry~\cite{gualtieri_pick_2018,kalashnikov_qt-opt_2018}, as it is considered to be one of the most important challenges for learning-based robotic grasping. In~\cite{kalashnikov_qt-opt_2018}, a general policy capable of grasping diverse objects was achieved by training on multiple real robots over the course of several weeks. Our system strives to achieve the same goal while training agents solely inside a simulation. To bridge a possible reality gap per our first contribution, we create a simulation environment with realistic physics and physically-based rendering while relying on domain randomization~\cite{tobin_domain_2017}, which is a popular technique to facilitate the sim-to-real transfer.

Manipulation tasks with high-dimensional continuous action and observation spaces pose many challenges when applying reinforcement learning due to its sample inefficiency, brittleness to hyperparameters, training instability and limited reproducibility~\cite{kroemer_review_2021}. Therefore, several different approaches have been analyzed over the years. Pixel-wise action space has been exploited for selecting action primitives based on traditional motion planning techniques in previous research such as~\cite{zeng_learning_2018}, where a grasp pose synthesis was incorporated with the pushing of objects. Alternative approaches focus on end-to-end learning to directly control the motion of robots either via joint commands~\cite{popov_data-efficient_2017,levine_end--end_2015} or actions that are expressed as Cartesian displacement of the gripper pose~\cite{kalashnikov_qt-opt_2018}. Our proposal similarly employs end-to-end learning of a policy that maps raw observations directly to continuous actions in Cartesian space.

The vast majority of research using deep reinforcement learning for robotic grasping relies on visual image observations coupled with convolutional neural networks. Here, RGB images are commonly used~\cite{tobin_domain_2017,kalashnikov_qt-opt_2018,levine_end--end_2015}, where depth maps or their inclusion in RGB-D data are also common~\cite{zeng_learning_2018}. However, it is argued that 2D convolutional layers do not provide the desired level of generalization over spatial position and orientation for robotic manipulation compared to their well-established generalization over the horizontal and vertical position in the image plane~\cite{gualtieri_pick_2018}. Therefore, our approach employs 3D octree observations due to advancements in their utilization in deep learning~\cite{wang_o-cnn_2017,riegler_octnet_2017}. As opposed to~\cite{wang_o-cnn_2017} that employs octree-based convolutional neural networks to analyze 3D shapes, we utilize octrees for real-time control with deep reinforcement learning. Although~\cite{trasnea_octopath_2021} recently applied octrees to learn planning of trajectories for mobile robots due to their efficiency, there is currently a lack of methods that employ 3D observations for end-to-end learning of manipulation skills. In this way, we study the importance of generalization over the full 6~DOF workspace in which robots operate and compare them to more traditional image-based observations as a part of our second contribution.

The application of this paper is closely related to the work presented in~\cite{grimm_vision-based_2021} and~\cite{wermelinger_grasping_2021}, both of which focus on grasping stones and boulders. Multi-finger humanoid hand is used in~\cite{grimm_vision-based_2021} with computationally-efficient control that enables both grasping and pushing actions. In~\cite{wermelinger_grasping_2021}, an excavator with a two-jaw gripper is used for the autonomous assembly of a large-scale stone wall with the capability to reorient grasped boulders. Whereas these approaches perform 3D object segmentation followed by grasp synthesis and traditional motion planning, our work focuses on end-to-end reinforcement learning. This enables agents to explore the necessary interactions with objects in a self-supervised manner, without the need to manually define complex subroutines such as pushing and reorientation. Furthermore, our focus on lunar environments introduces additional challenges such as uneven terrain and demanding illumination conditions.

%% file: content/methods.tex
\section{Problem Formulation}\label{sec:problem-formulation}

We focus on the robotic grasping of visually-perceived objects within the reachable workspace. It is assumed that any previous or subsequent motion of the rover and its arm is performed by traditional methods or other policies that are part of a larger hierarchy. We do not distinguish between isolated objects and those in cluttered scenes because rovers in extraterrestrial conditions would encounter both cases. We consider the task to be episodic, where each episode is successful once an object is grasped and lifted above a required threshold. Due to uneven terrain, this requirement is specified to be~25~cm above the base footprint of the robot. Each episode is further limited to~40~s, meaning that termination with failure occurs after a corresponding number of steps without success.

The task of robotic grasping can be formulated as a Markov decision process, where the behavior of an agent is defined by a policy~\mbox{\(\pi:\mathcal{S}\rightarrow\mathcal{A}\)} that provides a mapping from states to actions. At each discrete time step~\(t\), the environment utilizes a reward function~\mbox{\(r(s_{t},a_{t})\!\in\!\mathbb{R}\)} to emit an immediate reward for executing an action~\(a_{t}\) in a state~\(s_{t}\), which brings the agent to the next state~\(s_{t+1}\). The primary objective of the agent is to find the optimal policy~\(\pi^{*}\) that maximizes the expected sum of all future rewards~\mbox{\(\sum_{i=t}^T \gamma^{i-t}r(s_{i},a_{i})\)}, which are discounted by~\mbox{\(\gamma\!\in\![0, 1]\)} over a time horizon. Due to our focus on episodic grasping,~\(T\) indicates the terminal state of an episode.

\begin{figure*}[thpb]
	\vspace{1.379mm}
	\centering
	\includegraphics[width=1.0\textwidth]{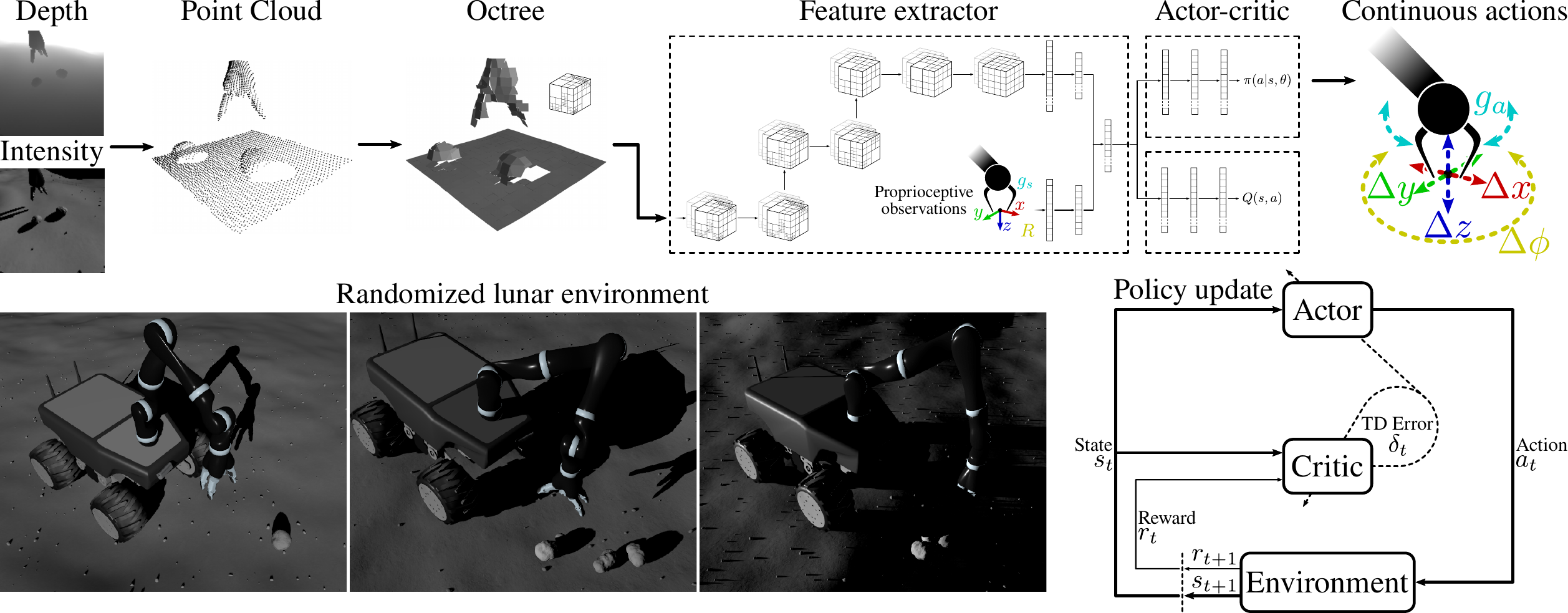}
	\caption{Overview of our approach for training reinforcement learning agents using a model-free actor-critic algorithm in lunar environments. For visual observations, we employ octrees that are constructed from depth maps and monochromatic images via an intermediate representation in the form of point clouds. Together with proprioceptive observations, a shared feature extractor is utilized to provide abstract features that are used as input for both actor and critic networks. The actor network optimizes a stochastic policy that provides continuous actions in Cartesian space in the form of gripper displacement and action. All of these networks are optimized simultaneously via end-to-end learning while an agent is trained inside a randomized simulation environment.}
	\label{fig:system-overview}
\end{figure*}

\section{Learning to Grasp from Octree Observations}\label{sec:learning-to-grasp-from-octree-observations}

Our approach incorporates visual octree observations with end-to-end learning of robotic grasping in a procedurally-generated simulation environment. An overview of the approach is presented in Fig.~\ref{fig:system-overview}.

\subsection{Truncated Quantile Critics}\label{ssec:truncated-quantile-critics}

We use Truncated Quantile Critics (TQC)~\cite{kuznetsov_controlling_2020} in this work to learn a policy for robotic grasping. It is a model-free off-policy actor-critic algorithm that incorporates both an actor for optimizing a stochastic policy~\mbox{\(\pi(a \vert s, \theta)\)} parameterized by \(\theta\), and one or more critics to estimate the action-value function~\mbox{\(Q(s, a)\)} by iteratively minimizing the temporal difference error \(\delta_{t}\) for a set of all available continuous actions that the agent is allowed to take. Critics evaluate the actor based on how rewarding the selected actions are. TQC is an extension of Soft Actor-Critic (SAC)~\cite{haarnoja_soft_2018} and therefore employs entropy regularization to optimize a trade-off between expected return and entropy, which represents the randomness of the policy. Unlike SAC, TQC utilizes a distributional representation for the action-value function~\mbox{\(Q(s, a)\)} of critics and truncates a number of topmost atoms from such distributions to address the problems with overestimation. We employ the specific implementation of TQC from Stable Baselines3~\cite{raffin_stable-baselines3_2021}.

\subsection{Observation Space}\label{ssec:observation-space}

The observation space of the agent consists primarily of visual observations that are represented as 3D octrees. Visual data originates from a stereo camera that is considered to be a space-suitable sensor as it is already used onboard current rovers~\cite{grotzinger_mars_2012}. In addition to depth maps produced by the camera, we also use monochromatic images captured by one of the imaging sensors to provide agents with supplementary information about the scene, e.g. the notion of shadows, as visualized in Fig.~\ref{fig:system-overview}. No meaningful information would be gained from multi-channel color images because lunar environments are generally homogeneously colored. Proprioceptive observations are also included to provide information about the state of the gripper in the case it is occluded or outside of the camera view. Only the gripper pose~\mbox{\((x,y,z,R_{1},R_{2},R_{3},R_{4},R_{5},R_{6})\)} and its state~\mbox{\(g_{s}\!\in\!\{closed\!:-1,open\!:1\}\)} are used in order to keep the observation space invariant to the utilized robot. Gripper pose is represented with respect to the robot base, and its orientation \(R\) is expressed as a continuous 6D rotation representation from~\cite{zhou_continuity_2019} due to its suitability for deep learning. Furthermore, the last two observations are stacked together at each time step, similar to the image stacking method presented in~\cite{mnih_human-level_2015}. This promotes the Markov property of the decision process formulation by providing the agent with temporal information about environment states and dynamics, such as the notion of motion.

Conversion of visual observations to octrees begins with the aforementioned depth map and monochromatic image, which are used to create a point cloud of the scene in the form of an unstructured list of~\mbox{\((x,y,z,i)\)} tuples. This intermediate representation is then transformed into the coordinate frame of the robot base. Such transformation between the camera and robot is assumed to be known or estimated via a hand-eye calibration procedure. Once transformed, the point cloud is cropped to occupy a fixed volume in space with a uniform aspect ratio. All remaining points are then used to construct an octree by performing a recursive subdivision of the occupied 3D volume until a certain depth~\(d_{max}\) is reached. Contrary to the inefficiency of voxel grids, octrees create hierarchical tree-like data structures where only occupied cells are recursively decomposed into eight child octants. Observable features can then be stored as channels at the finest leaf octants of the new octree, i.e.~the smallest possible octants at depth~\(d_{max}\). For each of these octants, all points from the point cloud that occupy the same volume in space are used to extract the relevant features. We store three distinct attributes at each finest leaf octant, i.e.~the average unit normal vector~\mbox{\((\overline{n}_{x},\overline{n}_{y},\overline{n}_{z})\)} estimated from the points, the average distance~\(\overline{d}\) from the center of an octant cell to all points used during its formation, and the average intensity~\(\overline{i}\). Both~\(\overline{d}\) and~\(\overline{i}\) are normalized to be in the range~\([0,1]\).

\subsection{Action Space}\label{ssec:action-space}

The agent is allowed to control the motion of the robotic arm at each time step through continuous actions in Cartesian space. These were selected in favor of joint commands due to their invariance to the specific kinematic configuration of the utilized robot. As illustrated in Fig.~\ref{fig:system-overview}, the action space for control of the gripper pose incorporates the relative translation~\mbox{\((\Delta{x},\Delta{y},\Delta{z})\!\in\![-1,1]\)} and yaw rotation~\mbox{\(\Delta{\phi}\!\in\![-1,1]\)}, which are expressed with respect to the robot base frame. Prior to executing these actions, they are mapped to their respective metric and angular ranges of \(\pm\)10~cm and \(\pm\)45\textdegree. Traditional collision-free motion planning and execution are accomplished by MoveIt~2~\cite{moveit2} while utilizing \mbox{TRAC-IK}~\cite{beeson_trac-ik_2015} kinematics solver and \mbox{RRTConnect}~\cite{kuffner_rrt-connect_2000} for path planning. Control of the gripper is similarly accomplished via a continuous action~\mbox{\(g_{a}\!\in\![-1,1]\)}, where positive values open and negative values close the gripper. The control frequency of the agent is set to 2.5~Hz because it is designed to provide high-level decision-making commands, while the lower-level controllers running at 200~Hz in simulation and 500~Hz on the real robot are responsible for real-time execution. With the aforementioned limitation of~40~s, this results in a maximum of~100 discrete time steps per episode during which the agent can select its actions.

\begin{figure*}[thpb]
	\vspace{1.379mm}
	\centering
	\includegraphics[width=1.0\textwidth]{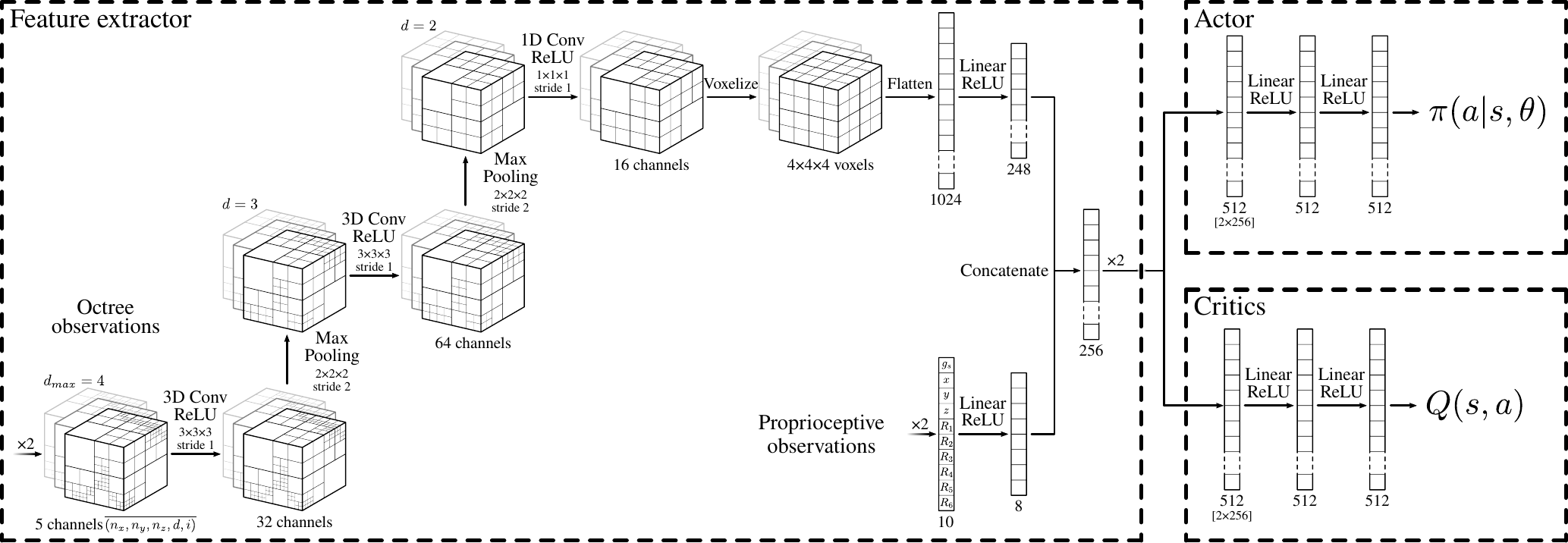}
	\caption{The full network architecture with a shared octree-based feature extractor and separate actor-critic networks.}
	\label{fig:actor-critic-network}
\end{figure*}

\subsection{Composite Reward Function}\label{ssec:composite-reward-function}

To accelerate training, we use a composite reward function that combines sparse rewards from four distinct stages of the entire grasping task, i.e.~reaching, touching, grasping and lifting. These follow a hierarchical flow, where each stage can give a reward upon its completion only once per episode. Their corresponding reward is set to increase exponentially as~\mbox{\(r_{b}^{r_{i}-1}\)} based on their order~\mbox{\(r_{i}\!\in\![1,4]\)} in the hierarchy. The base of the exponential function can be tuned, where~\mbox{\(r_{b}\!=\!8\)} was selected for TQC through empirical evaluation. This results in a theoretical maximum reward of~585 for each successful episode. In addition to the composite reward, a small reward of~\(-0.1\) is subtracted at each time step until the termination in order to encourage the agent to accomplish the task as fast as possible.

\subsection{End-to-End Learning from Octrees}\label{ssec:end-to-end-learning-from-octrees}

We employ a novel end-to-end approach for learning a policy that maps 3D octree observations directly into continuous actions via function approximation in the form of neural networks that are visualized in Fig.~\ref{fig:actor-critic-network}. Even though TQC requires separate networks for the actor and critics, it is beneficial to share a portion of their networks to reduce the total number of learnable parameters when dealing with high-dimensional visual observations such as octrees. Therefore, we adapt the octree-based convolutional neural network architecture from~\cite{wang_o-cnn_2017} into a feature extracting module that is optimized simultaneously with the actor and critic networks during the training. We utilize octrees with a maximum depth of~\mbox{\(d_{max}=4\)} as the input, which are processed through a series of 3D convolutional and pooling layers that progressively increase the number of channels while reducing the depth of the octree. In order to enable the use of standard network layers that operate on data structures with a fixed size, each octree is voxelized after reaching the depth~\mbox{\(d=2\)} by assigning zeros to all channels of every empty cell. Visual and proprioceptive features are then concatenated together into a feature vector. The feature extractor is applied in parallel to the entire temporally-stacked observation as a part of the same mini-batch. The output of this operation is then concatenated into a single feature vector that is used as input for the separate actor-critic networks that comprise of fully-connected layers.

\subsection{Training Environment}\label{ssec:training-environment}

We create a new simulation environment for the training of agents under lunar conditions. Our environment provides a standardized OpenAI Gym~\cite{brockman_openai_2016} interface while utilizing Gazebo~\cite{gazebo} robotics simulator (formerly known as Ignition). Gazebo was selected due to its open-source nature that encourages reproducibility and plugin-based architecture that makes it easily extensible. We employ DART physics engine with a step size of 5~ms to simulate rigid-body dynamics. Furthermore, physically-based rendering (PBR) capabilities are facilitated by using OGRE~2. Gym-Ignition~\cite{ferigo_gym-ignition_2020} is utilized for interfacing Gazebo due to its focus on reinforcement learning research. Lastly, ROS~2~\cite{ros2} is used as the middleware to facilitate communication among the primary nodes while simplifying the sim-to-real transfer by providing an identical interface for both simulated and real robots.

Datasets of models for terrain and objects are necessary to simulate a realistic-looking lunar environment that encapsulates its variety. However, there is a lack of available datasets from this domain that would provide the required level of detail at the scale of the robot workspace. For this reason, we create our datasets for both lunar surfaces and rocks through procedural generation. Our synthetic mesh generation pipelines are based on the Geometry Nodes feature of Blender~\cite{blender}. For each model, we apply displacement to individual vertices by a randomized 3D pattern that is formed as a combination of multiple random procedural textures at different scales. For lunar terrain, a subdivided 2D plane is used as the original geometry while imitating uneven terrain with impact craters. Lunar rocks are initialized from subdivided convex polyhedrons that are displaced to represent rocks of different sizes and shapes, including non-convex geometry. To improve training times, each model is generated at two different levels of detail, where a simpler one is used for its collision geometry. Furthermore, each model is assigned a random set of PBR material textures during its insertion into the environment. Using this simple technique, we can synthesize unique environments with nearly an unlimited number of permutations. During our experiments, we utilize a dataset with 250~surfaces, 250~rocks and 35~PBR texture sets.

Domain randomization is employed in order to increase the variety of the environment even further, which is illustrated by the initial states of three example episodes in Fig.~\ref{fig:system-overview}. The following attributes are randomized from uniform distribution during environment resets:~the pose of the rover within the environment and the initial joint configuration of its arm above the workspace; terrain model, its surface friction and material; object count~(1~--~4), their models, densities, surface frictions, materials and poses relative to the rover; as well as direction and elevation of the simulated Sun. The pose of the virtual camera is also randomized relative to its mounting surface on the rover~(\(\pm\)5 cm). Furthermore, Gaussian noise~\mbox{\(\mathcal{N}(0, 0.001)\)} is added to the captured depth maps and monochromatic images in order to increase the realism of observations.

\subsection{Curriculum}\label{ssec:curriculum}

In order to accelerate the training in the early stages, a curriculum is applied to the height requirement for the successful lifting of objects. This requirement is set to be proportional to the current success rate smoothed by a rolling average over the last~100 episodes. The requirement is initialized at 7.5~cm and increased linearly to 25~cm until a success rate of 33\% is reached. With this addition, successful grasps become more common during the early stages of training while the actions of agents are nearly random.

%% file: content/results.tex
\section{Experiments}\label{sec:experiments}

We evaluated the real-world applicability of our approach while comparing the 2D image and 3D octree observations.

\subsection{Experimental Setup}\label{ssec:experimental-setup}

Throughout the experiments, we used a Summit XL-GEN mobile manipulator from Robotnik equipped with a 7~DOF Kinova Gen2 robotic arm and a three-finger mechanical gripper. The robot was controlled via MoveIt~2 both inside simulation for training and during sim-to-real evaluation in a Moon-analogue facility. To obtain visual observations inside the simulation, a virtual camera with a resolution of 128\(\times\)128~px was used at a framerate of 15~Hz. On the real robot, we used Intel RealSense D435 to capture images of 424\(\times\)240~px at 15~Hz, which were cropped and resized to 128\(\times\)128~px before use. We attached the camera statically at the front of the rover while randomizing its simulated pose during the training. To estimate the transformation between the real camera and the base of the real robot, we performed a hand-eye calibration. We did not employ any additional artificial lights mounted on the robot and relied solely on ambient illumination that originated either from the simulated Sun or a light source inside the Moon-analogue facility that emulates the solar illumination. For octrees, we used an observable volume of~40\(\times\)40\(\times\)40~cm in front of the rover, which results in the size of~2.5~cm for the finest leaf octants at the selected maximum depth of~\(d_{max}=4\). In order to avoid exploring areas of no interest, the position of the gripper was restricted to a~35\(\times\)35\(\times\)60~cm workspace centered at the origin of observable volume.

\subsection{Hyperparameters}\label{ssec:hyperparameters}

Because the selection of hyperparameters can significantly affect the learning curve and final performance of learned policies, we optimized hyperparameters in two consecutive steps. First, an automatic optimization was performed using the hyperparameter optimization framework Optuna~\cite{akiba_optuna_2019}. This was followed by subsequent fine-tuning of targeted hyperparameters via manual optimization. During this process, we also tuned the aforementioned reward scale, the number of observation stacks and the size of neural networks. The utilized set of hyperparameters can be seen in Table~\ref{tab:hyperaparameters}.

\begin{table}[ht]
	\centering
	\caption{Hyperparameters used for the training of all agents.}
	\label{tab:hyperaparameters}
	\vspace{-0.25\baselineskip}
	\begin{tabular}{lc}
		\hline
		\textbf{Hyperparameter}        & \textbf{TQC}                                \\ \hline
		Optimization algorithm         & Adam~\cite{kingma_adam_2014}                \\
		Learning rate schedule         & Linear, \(2.0 \cdot 10^{-4} \rightarrow 0\) \\
		Mini-batch size                & \(64\)                                      \\
		Gradient steps per update      & \(100\) (after every episode)               \\
		Size of the replay buffer      & \(50000\)                                   \\
		Discount factor \(\gamma\)     & \(0.99\)                                    \\
		Target update rate \(\tau\)    & \(4 \cdot 10^{-5}\)                         \\
		Entropy coefficient \(\alpha\) & Automatic~\cite{haarnoja_soft_2018}         \\
		Entropy target                 & \(-dim(\mathcal{A})=-5\)                    \\
		Number of critics              & \(2\)                                       \\
		Number of atoms                & \(25\)                                      \\
		Number of truncated atoms      & \(3\)                                       \\
		Exploratory action noise       & \(\mathcal{N}(0, 0.025)\)                   \\  \hline
	\end{tabular}
	\vspace{-0.8\baselineskip}
\end{table}

\subsection{Training Procedure}\label{ssec:training-procedure}

\begin{figure}[ht]
	\centering
	\includegraphics[width=1.0\linewidth]{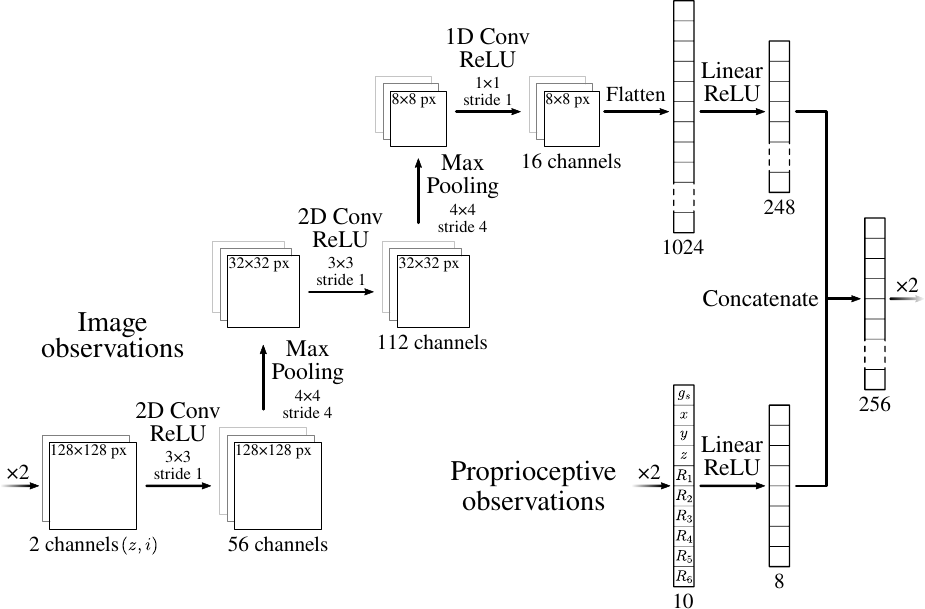}
	\caption{The network architecture of the image-based feature extractor, which is analogous to the octree-based feature extractor presented in Fig.~\ref{fig:actor-critic-network}.}
	\label{fig:image-feature-extractor}
\end{figure}

We trained agents in our simulation environment to learn policies for grasping lunar rocks by the use of TQC. In order to evaluate the proposed use of octree observations, we trained separate agents that use either image or octree observations. For image observations, we employed images with the resolution of 128\(\times\)128~px and two channels that contain the depth and intensity values for each pixel. Both of these values are normalized to be in the range~\([0,1]\), where the maximum distance of~1~m is selected for the depth map. Similar to octrees, two consecutive images are stacked together for each observation to preserve temporal information. To provide a fair comparison, an analogous feature extraction network architecture is created for image observations based on the octree-based feature extractor from Fig.~\ref{fig:actor-critic-network} by replacing 3D operations with their 2D variants. To achieve approximately the same number of learnable parameters for the feature extractor, i.e.~315k parameters, we increase the number of channels in 2D convolutional layers. The resulting network architecture for the image-based feature extractor is illustrated in Fig.~\ref{fig:image-feature-extractor}.

We also evaluated the importance of domain randomization for sim-to-real transfer in lunar applications. To do this, we trained agents on two different variants of the environment for each observation type. The first variant is the environment with complete domain randomization, whereas the second variant has a reduced level of randomization. In this reduced environment, only the pose of objects and the initial joint configuration are randomized for each episode. From the utilized datasets, one model of the lunar surface and four models of lunar rocks are randomly selected and used throughout the training. All other variables, such as the pose of the camera and the direction of illumination, are similarly selected at random but kept static throughout the training.

\begin{figure}[ht]
	\centering
	\includegraphics[width=1.0\linewidth]{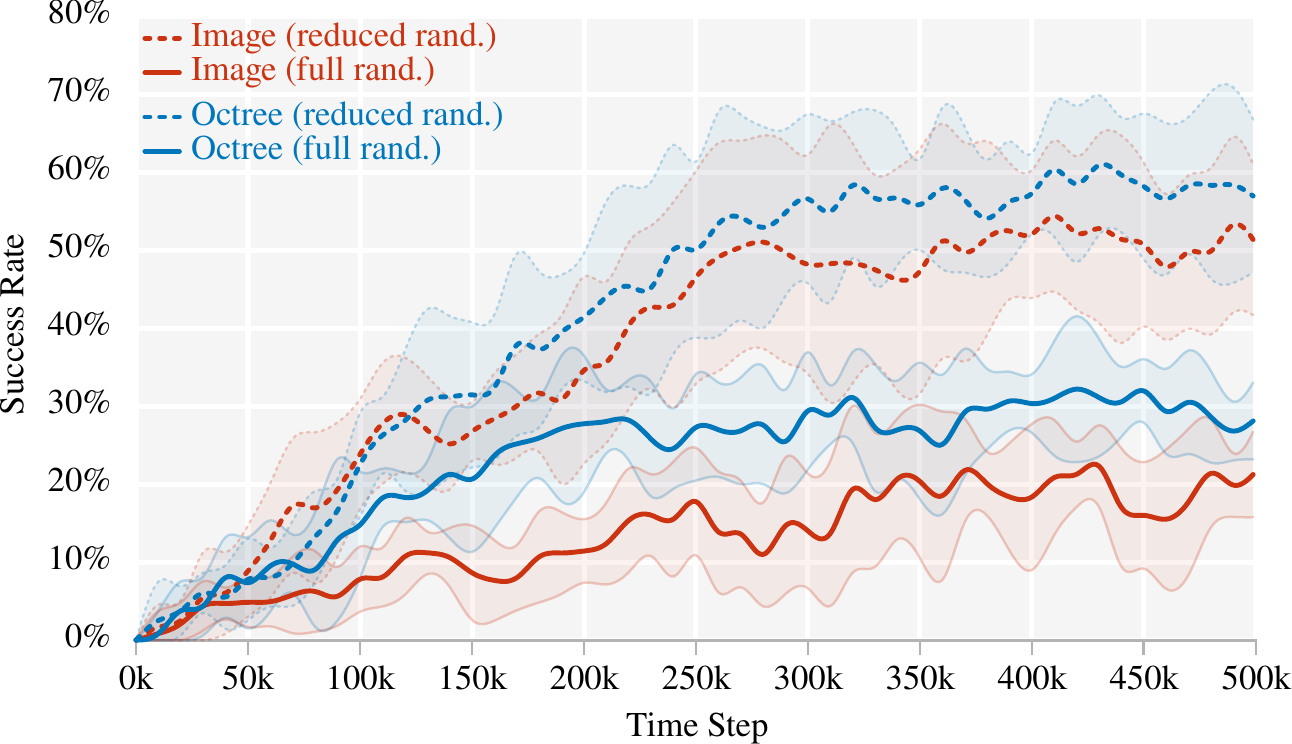}
	\caption{Learning curves of the trained agents that are evaluated every 10k steps. Solid lines represent smoothed mean, whereas shaded areas indicate the standard deviation over three runs with different random seeds.}
	\label{fig:results-learning-curve}
\end{figure}

All agents were trained for 500k time steps and periodically evaluated every 10k steps on 20 episodes by utilizing their current policies with deterministic actions. For each analyzed variant, we trained three agents with different seeds for the pseudorandom generator that initializes the simulation environment and all learnable parameters. Since we evaluate the agents in a Moon-analogue facility located on Earth, the gravity of the simulated environment is set to~9.807~m/s\(^2\). Learning curves of all agents can be seen in Fig.~\ref{fig:results-learning-curve}.

\subsection{Sim-to-Real Transfer}\label{ssec:sim-to-real-transfer}

\begin{figure}[ht]
	\centering
	\includegraphics[width=0.8\linewidth]{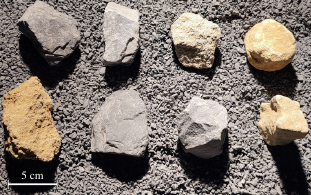}
	\caption{Eight rocks used during the evaluation of the sim-to-real transfer.}
	\label{fig:test-rocks}
\end{figure}

We evaluated the feasibility of sim-to-real transfer in LunaLab~\cite{ludivig_building_2020}, which is a Moon-analogue facility at the University of Luxembourg. We assessed all variants of trained agents, where the best performing agent at the end of its training inside the simulation is selected for each variant. The policies of these agents were then used for deterministic selection of actions during execution on the real robot. A~total of eight different rocks from Fig.~\ref{fig:test-rocks} were utilized during this experiment, where~1~--~4 rocks were randomly scattered within the workspace for each episode. Every agent was evaluated over the course of 25 episodes that were considered successful if the robot grasped and lifted one of the rocks.

\begin{figure}[ht]
	\centering
	\includegraphics[width=1.0\linewidth]{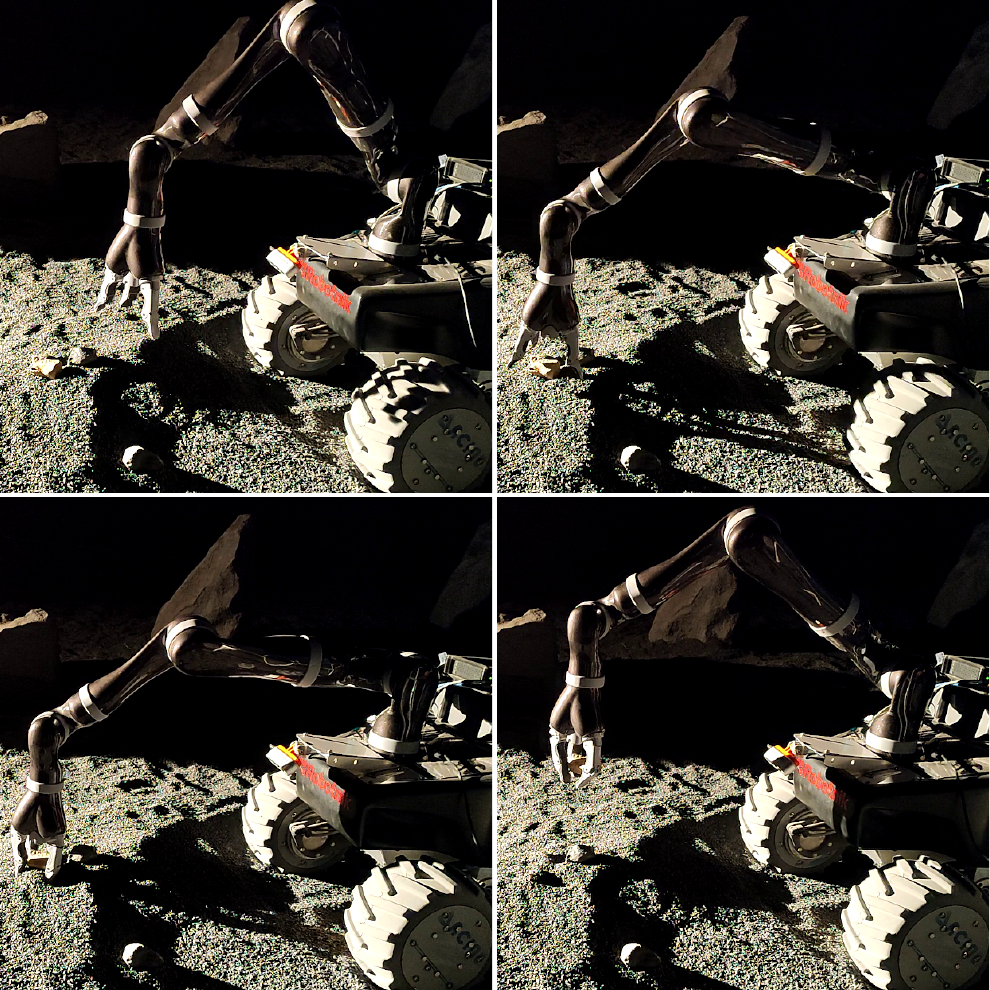}
	\caption{An example of a successful grasp sequence using the real robot.}
	\label{fig:grasp-sequence}
	\vspace{-1.0\baselineskip}
\end{figure}

Quantitative results of the sim-to-real success rate appear in Table~\ref{tab:sim_to_real_success_rate}, whereas an example of a successful grasp sequence is visualized in Fig.~\ref{fig:grasp-sequence}. During episodes that failed due to timeout, it was observed that agents got frequently stuck in a loop attempting to repeatedly grasp a specific rock without success due to improper positioning of the gripper. This behavior occurred for all evaluated agent variants, but it was most prominent for agents trained in environments with reduced domain randomization.

\begin{table}[ht]
	\vspace{1.379mm}
	\centering
	\caption{Quantitative results of the zero-shot sim-to-real transfer.}
	\label{tab:sim_to_real_success_rate}
	\vspace{-0.25\baselineskip}
	\begin{tabular}{ccc}
		\hline
		\textbf{Observation type} & \textbf{Level of randomization} & \textbf{Success rate} (n=25) \\ \hline
		Image                     & Reduced                         & 12\%                         \\
		Image                     & Full                            & 20\%                         \\
		Octree                    & Reduced                         & {\phantom{0}}8\%             \\
		Octree                    & Full                            & 32\%                         \\ \hline
	\end{tabular}
	\vspace{-0.8\baselineskip}
\end{table}

%% file: content/discussion_and_conclusion.tex
\section{Discussion}\label{sec:discussion}

By training agents at two different levels of domain randomization, we investigated its effect on the sim-to-real transfer. However, the level of randomization significantly impacted the overall learning of agents. With fully-enabled randomization, agents encountered training instability that restricted their achievable success rate compared to agents that experienced a reduced subset of the lunar variety. We attribute the instability to high variance in gradient estimation due to large variations in observations. Despite this, sim-to-real experiments indicate that domain randomization enables robust transferability to the real-world domain, where agents trained in fully randomized environments performed better when transferred to the real robot despite their lower success rate in simulation. Although domain randomization enables learning of policies capable of generalization under challenging conditions, training stability must be significantly improved via methods such as active domain randomization~\cite{mehta_2019_active} in order to take full advantage of this approach.

The experimental evaluation indicates that 3D visual observations in the form of octrees provide better performance than image-based observations when applied for end-to-end learning of robotic grasping. This result is attributed to the better ability of 3D convolutions to generalize over spatial positions and orientations, unlike 2D convolutions that generalize well over planar image coordinates. Another advantage of 3D observations is their ability to provide learned policies with invariance to the camera pose, further simplifying the transfer to a different system or application domain. Sensory fusion of data from multiple depth sensors could also be applied to obtain a single 3D data structure that provides observations with improved quality and reduced occlusions. Therefore, these and other techniques must be investigated further to reveal the full potential of 3D observations in self-supervised learning of robot manipulation skills.

\section{Conclusion}\label{sec:conclusion}

In this work, we presented an approach for learning robotic grasping on the Moon with end-to-end deep reinforcement learning. We analyzed the application of 3D octree observations and compared their performance to 2D images. We also investigated the effects of employing domain randomization in lunar environments by demonstrating zero-shot sim-to-real transfer to a real robot in a Moon-analogue facility. Overall, we believe that deep reinforcement learning is a promising method for acquiring various manipulation skills for robots in space, despite its many challenges. Improving the learning stability in diverse environments is one of the necessary steps before similar approaches can be robustly employed for the wide range of applications within space robotics.